\documentclass[a4paper,fleqn]{cas-dc}

\usepackage[numbers,sort,compress]{natbib}

\def\tsc#1{\csdef{#1}{\textsc{\lowercase{#1}}\xspace}}
\tsc{WGM}
\tsc{QE}

\usepackage{multirow}
\usepackage{makecell}
\usepackage{stfloats}
\usepackage{bbding}
\usepackage{amsmath}
\usepackage{algorithmic}
\usepackage{algorithm,algorithmic}
\usepackage{graphics}

\begin{document}
\let\WriteBookmarks\relax
\def\floatpagepagefraction{1}
\def\textpagefraction{.001}
\let\printorcid\relax

\shorttitle{SELECTOR: Heterogeneous graph network with convolutional masked autoencoder for multimodal robust prediction of cancer survival}    

\shortauthors{L. Pan et al.}  

\title [mode = title]{SELECTOR: Heterogeneous graph network with convolutional masked autoencoder for multimodal robust prediction of cancer survival}  

\author[a]{Liangrui Pan} \ead{panlr@hnu.edu.cn}
\author[a]{Yijun Peng} \ead{pengyijun@hnu.edu.cn}
\author[a]{Yan Li} \ead{s2310w1062@hnu.edu.cn}
\author[b]{Xiang Wang} \ead{wangxiang@csu.edu.cn}
\author[a]{Wenjuan Liu} \ead{liuwenjuan89@hnu.edu }
\author[a]{Liwen Xu} \cormark[1]\ead{xuliwen@hnu.edu.cn}
\author[c]{Qingchun Liang} \cormark[1] \ead{503079@csu.edu.cn}
\author[a]{Shaoliang Peng}\cormark[1]\ead{slpeng@hnu.edu.cn}

\affiliation[a]{organization={College of Computer Science and Electronic Engineering, Hunan University},
				city={Changsha},
				postcode={410083}, 
				state={Hunan},
				country={China}}

\affiliation[b]{organization={Department of Thoracic Surgery, The second xiangya hospital, Central South University},
	city={Changsha},
	postcode={410083}, 
	state={Hunan},
	country={China}}

\affiliation[c]{organization={Department of Pathology, The second xiangya hospital, Central South University},
	city={Changsha},
	postcode={410083}, 
	state={Hunan},
	country={China}}
\cortext[1]{Corresponding author}

\begin{abstract}
Accurately predicting the survival rate of cancer patients is crucial for aiding clinicians in planning appropriate treatment, reducing cancer-related medical expenses, and significantly enhancing patients' quality of life. Multimodal prediction of cancer patient survival offers a more comprehensive and precise approach. However, existing methods still grapple with challenges related to missing multimodal data and information interaction within modalities. This paper introduces SELECTOR, a heterogeneous graph-aware network based on convolutional mask encoders for robust multimodal prediction of cancer patient survival. SELECTOR comprises feature edge reconstruction, convolutional mask encoder, feature cross-fusion, and multimodal survival prediction modules. Initially, we construct a multimodal heterogeneous graph and employ the meta-path method for feature edge reconstruction, ensuring comprehensive incorporation of feature information from graph edges and effective embedding of nodes. To mitigate the impact of missing features within the modality on prediction accuracy, we devised a convolutional masked autoencoder (CMAE) to process the heterogeneous graph post-feature reconstruction. Subsequently, the feature cross-fusion module facilitates communication between modalities, ensuring that output features encompass all features of the modality and relevant information from other modalities. Extensive experiments and analysis on six cancer datasets from TCGA demonstrate that our method significantly outperforms state-of-the-art methods in both modality-missing and intra-modality information-confirmed cases. Our codes are made available at \url{https://github.com/panliangrui/Selector}.
\end{abstract}



\maketitle

\section{Introduction}
\label{sec_introduction}
Cancer is a globally significant health challenge \cite{williams2023aacr}, despite remarkable strides in its combat. The worldwide prevalence of cancer continues to pose a substantial public health challenge, exacerbated by factors such as global population growth, aging, and the recent COVID-19 pandemic \cite{brenner2022vitamin}. The complexity of cancer, variations in pathological and molecular characteristics among patients, and diverse clinical profiles present substantial barriers to tailored treatment and precise prognosis \cite{pan2023multi,garcia2015missing}. Accurate survival prediction for cancer patients is pivotal in evaluating prognosis, facilitating collaborative decision-making between patients and clinicians, and devising personalized treatment plans.

Cancer survival prediction assesses and forecasts the survival of cancer patients using clinical record data, pathology data, and molecular feature data \cite{tran2021deep,pan2023pacs}. Its primary objective is to predict the patient's future survival probability and time-based on individual characteristics and disease information. In recent years, various methods have been proposed for cancer survival prediction, primarily leveraging multimodal data, including histopathology images, omics data (genome, copy number variation, methylation, etc.), and clinical records \cite{deng2021integrating,he2023artificial}. Pathological images offer detailed information on tumor tissue, aiding in tumor type and grade determination and revealing histological characteristics and molecular marker expression \cite{yao2020whole}. Hence, pathology images serve as the gold standard for cancer diagnosis \cite{di2022big,pan2022noise,pan2023ldcsf}. Cancer is intimately linked to genomic mutations or abnormal gene expression altering normal cell functions and biological processes—essentially, genes determine the phenotype \cite{liang2020novel}. Genomic data plays a crucial role in cancer survival prediction. However, genome-related data, such as copy number variation and methylation, also reflect genetic characteristic changes, unveiling the molecular mechanisms of cancer, disease subtypes, and potential therapeutic targets. Clinical records encompass patient basic information, medical history, follow-up processes, treatment plans, etc., serving as vital indicators for evaluating patient survival prediction \cite{liu2020optimizing,chen2014risk}.

Given the significance of multimodal data in cancer, multimodality can offer more comprehensive and accurate information for disease diagnosis, staging, treatment regimen selection, prognostic assessment, risk prediction, early detection, and recurrence monitoring. This contributes to enhancing the survival prediction of patients and facilitating more individualized treatment strategies. Over the past few years, deep learning has found extensive application in the survival prediction of cancer patients \cite{steyaert2023multimodal,wu2023camr,wang2021gpdbn,li2022hfbsurv,fu2023deep}. The novelty in most of these works lies primarily in the methods of multimodal data feature extraction and multimodal data fusion. In the preprocessing of whole slide images (WSI), the first step involves identifying the valid portions of all organizational features in the WSI. These features are sliced, and each WSI contains numerous patches of size (224$\times$224, or 256$\times$256). Subsequently, pre-trained models such as Resnet50, Swim Transformer, and KimiaNet are utilized for the initial WSI feature extraction \cite{lu2021data,chen2022pan,hou2023hybrid}. Regarding the preprocessing of cancer modal data, the main steps involve feature selection, data transformation, and data normalization. SURVPATH proposes tagging transcriptomics with semantically meaningful, interpretable, and end-to-end learnable biological pathway markers \cite{jaume2023modeling}. HGCN suggests using Gene Set Enrichment Analysis (GSEA) to classify gene expression data into five groups \cite{hou2023hybrid}. Preprocessing clinical data typically involves outlier handling and normalization operations.

After preprocessing multimodal data, various machine learning algorithms and deep learning methods are employed for feature extraction. Common feature extraction methods encompass attention-based multiple instance learning (AMIL), self-normalizing network (SNN), graph convolutional network (GCN), etc \cite{ilse2018attention,hou2023hybrid}. Enhancing the accuracy of survival prediction is contingent upon the feature fusion method. The primary feature fusion methods consist of early fusion and late fusion. Early fusion methods are relatively scarce, with examples like multimodal co-attention transformer (MCAT) \cite{chen2021multimodal}. Late fusion methods abound, including attention-based multiple instance learning (AB-MIL) \cite{ilse2018attention}, GPDBN \cite{wang2021gpdbn}, deep attention multiple instance survival learning (DeepAttnMISL) \cite{yao2020whole}, transformer-based MIL (TransMIL) \cite{shao2021transmil}, and Pathomics \cite{chen2020pathomic}. Late fusion predominantly aims to improve the feature representation of each independent modality, utilizing relatively straightforward fusion schemes like concatenation, row vector maximum, outer product, similarity constraints, co-attention fusion, and other methods \cite{chen2021multimodal,chen2020pathomic,mobadersany2018predicting,vale2021long,cheerla2019deep}. Nevertheless, these methods fall short in facilitating interactions among data features within the same modality and across different modalities, thereby constraining the benefits of multimodal data fusion and diminishing the accuracy of survival prediction. Next, the lack of multimodal data is often due to patients' unwillingness to bear high medical expenses or doctors' negligence in comprehensively collecting multimodal data.

This paper proposes SELECTOR, a heterogeneous graph-aware network based on convolutional mask encoders for robust multimodal prediction of cancer patient survival. Leveraging the characteristics of multimodal heterogeneous graphs, SELECTOR introduces a feature edge reconstruction module to acquire effective node embeddings and edge features. It fully considers the semantic information within the intricate structure of heterogeneous graphs. Drawing inspiration from the ability of MAE to recover the original mask portion from features with missing details, enhancing model robustness, we introduce a convolutional mask autoencoder (CMAE). CMAE processes heterogeneous graphs after feature reconstruction to prevent missing information within the modality from impacting the accuracy of the prediction model. CMAE incorporates transformer training concepts, replacing all convolutional layers in the encoder part with sub-manifold sparse convolution. This design ensures the model operates solely on visible data features \cite{graham20183d}. Additionally, to foster interaction among feature information within multimodal modes, enriching information expression, learning complementary features, and mitigating noise and deletions, we introduce a feature interaction module to facilitate communication between multimodal modes. Ultimately, SELECTOR can predict patient survival in the presence of missing modal data or missing information within a modality. In summary, our main contributions encompass: 


\begin{itemize}
	\item [1)] Construct a multimodal heterogeneous graph and introduce the concept of a meta-path for the edge reconstruction of features. This ensures a comprehensive consideration of feature information at the graph edge and effective node embedding.
	\item [2)] To mitigate the impact of missing features within the modality on prediction accuracy, we introduce a CMAE for processing heterogeneous graphs after feature reconstruction. This method is primarily based on the concept of sparse convolution in feature extraction, extracting only unmasked features. 
	\item [3)] We design a feature cross-communication module to facilitate communication between multiple modalities. This ensures that the output features incorporate all modality-specific features along with relevant information from other modalities.
	\item [4)] SELECTOR can effectively predict cancer patient survival in the presence of missing intra-modal and modal data. Extensive experimental verification demonstrates that this framework achieves the highest prediction accuracy across six cancer multimodal datasets from TCGA.
\end{itemize}


\section{Related Work}
\subsection{Multimodal Cancer Survival Prediction}
In recent decades, specific survival predictions have relied on a combination of clinical experience and professional judgment. The advent of artificial intelligence has seen the application of deep learning methods for survival prediction. These methods enhance prediction performance by learning intricate feature representations and patterns from extensive medical images, genomics, and clinical data. In recent years, efforts to enhance the survival prediction accuracy for cancer patients have led to studies attempting to integrate pathology data, genomics, transcriptomics, and clinical data. This has spurred innovation in methods for multimodal survival prediction in cancer. The majority of these methods employ late fusion strategies, processing modalities differently. Fusion methods include vector splicing \cite{mobadersany2018predicting}, modal data aggregation \cite{cheerla2019deep}, kronecker product \cite{wang2021gpdbn,chen2020pathomic}, and factored bilinear pooling \cite{li2022hfbsurv}. For instance, cross-aligned multimodal representation learning (CAMR) introduces a cross-modal representation alignment learning network to bridge the modal gap. It effectively learns modality-invariant representations in a shared subspace and employs a cross-modal fusion module to integrate the modalities. State-invariant representations are integrated into a unified cross-modal representation for each patient \cite{wu2023camr}. Nevertheless, some works also delve into early fusion strategies, modeling cross-modal interactions in input data from various modalities. The MCAT learns an interpretable, dense co-attention mapping between WSI and genomic features formulated in embedding space \cite{chen2021multimodal}. Additionally, the framework comprehends how histological patches concentrate on genes when predicting patient survival \cite{chen2021multimodal}. While these methods have consistently improved the accuracy of multimodal cancer survival prediction, they overlook the challenge of missing multimodal data. This arises when patients cannot undergo physical examinations for various reasons or are unwilling to incur additional examination costs.

\subsection{Heterogeneous graph convolutional neural network}
Heterogeneous graphs are prevalent in various domains, simulating diverse relationships between different types of nodes, including food graphs, social graphs, biomedical graphs, etc. \cite{tian2023heterogeneous}. Consequently, researchers have introduced heterogeneous graph neural networks (HGNNs) to capture the intricate structural and semantic information within heterogeneous graphs. Random walk strategy is incorporated into HGNNs to sample fixed-size, strongly correlated heterogeneous neighbors for each node. These neighbors are grouped based on node type, and neural networks (NN) are subsequently employed to aggregate information from these sampled adjacent nodes \cite{zhang2019heterogeneous}. However, this method requires node labels for learning and cannot extract supervised signals from the data to learn general node embeddings \cite{tian2023heterogeneous}. Additionally, a novel framework, HGSL, can simultaneously address the classification tasks of heterogeneous graph structure learning and GNN parameter learning \cite{zhao2021heterogeneous}. Within each generated relation subgraph, HGSL not only considers feature similarity by producing feature similarity graphs but also takes into account intricate heterogeneous interactions in features and semantics through the generation of feature propagation graphs and semantic graphs. Nevertheless, most HGNNs depend on labeled data to guide semi-supervised or self-supervised learning. Hence, the efficient utilization of rich semantic information in heterogeneous graphs as a guide for self-supervised learning paradigms remains an unresolved challenge.

\subsection{Masked autoencoder}
The fundamental principle behind masked autoencoder (MAE) involves removing a portion of the data and training the model to predict the content of the deleted class \cite{he2022masked}. MAE appears to function more like a typical denoising autoencoder, a technique widely applied in natural language processing (e.g., ChatGPT), computer vision, and various other domains \cite{vincent2008extracting,thorp2023chatgpt}. To prevent input downsampling while ensuring computational efficiency, a patch-based autoencoder PVQVAE was devised. The encoder transforms masked images into non-overlapping patch tokens, and the decoder restores masked regions from painted tokens, while preserving the unmasked area unaltered \cite{liu2022reduce}. MAE employs a computationally efficient knowledge distillation framework, leveraging pre-trained models to robustly extract knowledge from graphs \cite{bai2023masked}. Contrastive mask autoencoder meticulously integrates contrastive learning and mask image models through innovative design. It leverages the strengths of different modules to acquire representations with robust instance discrimination and local awareness \cite{huang2023contrastive}. Drawing inspiration from prior generative models, MAGE incorporates semantic tags learned by vector quantized GANs into both input and output, combining them with masks \cite{li2023mage}. Thus, the MAE-based method effectively addresses the issue of missing features and can be applied to other datasets as well. Consequently, this article will investigate the utilization of MAE-related methods to address the challenge of missing multimodal data. 

\section{Methods}

\subsection{Overview}
\begin{figure*}[!t]
	\centerline{\includegraphics[width=2\columnwidth]{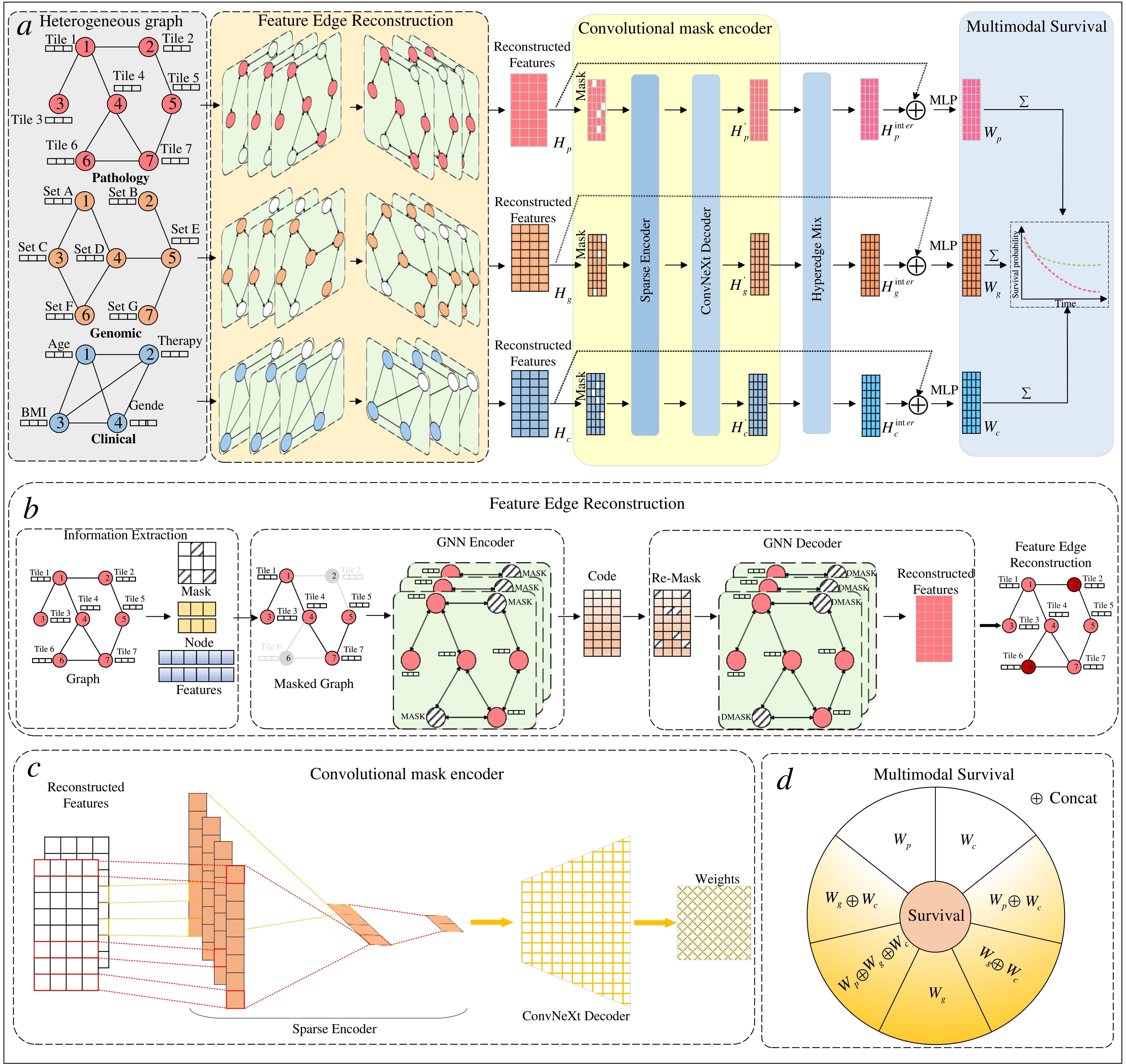}}
	\caption{ $a$: Flowchart of learning cancer multimodal datasets to predict patient survival using a heterogeneous graph-aware network based on generative encoders. $b$: Flowchart of feature edge reconstruction module based on multimodal graph. $c$: Workflow of fully convolution-based masked autoencoder. $d$: 7 schemes for multimodal prediction of patient survival.}
	\label{fig:fram}
\end{figure*}

Illustrated in Figure.~\ref{fig:fram}($a$), SELECTOR contains the construction of multimodal heterogeneous graphs, feature edge reconstruction, convolutional mask encoder, feature cross-fusion, and multimodal survival prediction modules. These modules collaborate to make survival predictions for cancer patients, considering the absence of modalities and intra-modal information.

\subsection{Construction of multimodal heterogeneous graphs}
To predict the survival of cancer patients using multimodal data, this experiment curated a set of pathological data, genetic data, and clinical records. These were selected to provide a comprehensive depiction of the patient's physical condition, spanning from the cellular to the molecular level. Let $D = \{ p,g,c\}$represent the multimodal data of the patient, with its observation time and examination status denoted as $({t_i},{\delta _i})$.

Given the relatively large size of WSI pixels, we begin by removing the background, preserving only the meaningful organizational elements. Subsequently, the tissue section is partitioned into 256$\times $256-sized blocks using a sliding block method. The pre-trained model KimiaNet is then employed to extract features for each block \cite{riasatian2021fine}, generating 1024-dimensional vectorized representations based on spatial location information. These representations are interconnected in an 8-adjacent manner to form a heterogeneous graphical representation of the pathology slice, denoted as ${G_p} = \{ {V_p},{E_p}\}$. This graph-based representation enhances the accuracy and efficiency in illustrating the pathology tissue topology, reducing computational load during model training and preprocessing for complex pathology studies. For constructing the genome graph, we chose 1742 genes from the genetic graph and classified them into five groups using Gene Set Enrichment Analysis (GSEA): tumor suppressors, oncogenes, cell differentiation markers, and cytokines and growth factors. Due to the interaction and connection among these gene groups, the resulting genome graph is represented heterogeneously as ${G_g} = \{ {V_g},{E_g}\}$. To create a graphical representation of clinical records, we quantified each record, including age, gender, BMI, etc., and applied one-hot encoding to generate a vector representation for each patient. By fully connecting the one-hot vectors, we obtain the heterogeneous graph representation of clinical records, denoted as ${G_c} = \{ {V_c},{E_c}\}$. In the multimodal graph representation, ${V_p},{V_g},{V_c}$ are represented as a feature set of nodes respectively. ${E_p},{E_g},{E_c}$ are respectively represented as a set of edges constructed based on neighborhood relationships. To unify the feature dimensions of multimodal data, we use zero padding to feature align all features that do not satisfy 1024 dimensions. As shown in Figure.~\ref{fig:fram}($a$), the multimodal heterogeneous graph we constructed is on the left side of the figure.

\subsection{Feature edge reconstruction}
In comparison to CNN feature extraction, graph convolution (GCN) takes into account both local features and the global topology of nodes, resulting in more comprehensive features. Additionally, it can adapt to changing graph structures, allowing nodes to be added or deleted during training. This adaptability enables the generation of meaningful feature representations and achieves optimal feature extraction results with fewer model parameters. For instance, clinical record data often contains missing features, primarily due to reasons such as sensor damage, data loss, and human recording errors. The flexibility of graph representation and inductive GCN allows patients with missing information to be included in relevant studies without disrupting the network's training and testing processes. This flexibility is challenging to achieve with fully connected layers. However, different nodes in the graph representation correspond to various types of attributes, implying distinct structural roles. To effectively learn node embeddings and comprehensively consider semantic information within the intricate structure of heterogeneous graphs, it is crucial to reconstruct the neglected features from feature edges.

Let's consider designing a heterogeneous graph, denoted as $G = \{ V,E\}$, composed of an object set and a connection set. The heterogeneous graph incorporates a node type mapping function $\phi :V \to A$ and a connection type mapping function $\psi :E \to B$, where $A$ and $B$ represent predefined object types and connection types, respectively, with $\left| A \right| + \left| B \right| > 2$ being their union. In a heterogeneous graph, two objects can be connected through various semantic paths, referred to as the meta-path $\psi $, defined as ${A_1}\mathop  \to \limits^{{R_1}} {A_2}\mathop  \to \limits^{{R_2}} ....\mathop  \to \limits^{{R_l}} A{}_{l + 1}$. This meta-path describes the composite relationship ${R_1} \circ {R_2} \circ ... \circ {R_{l + 1}}$ between objects ${A_1} \to {A_L}$, where $l$ is the path length, and $\circ $ represents the composite operation on the relationship. To obtain the feature edge reconstruction graph, we initially design an encoder L capable of learning node embeddings $H = {f_E}(G)$ to encode complex information in the heterogeneous graph. Subsequently, a decoder $M$ is devised to reconstruct the input features, denoted as ${G^{'}} = {f_D}(H)$. ${G^{'}}$ denotes the adjacency matrix of feature edge reconstruction.

In Figure.~\ref{fig:fram}($b$), we introduce a meta-path-based strategy for feature edge reconstruction (FER) to capture rich semantic information within heterogeneous graph structures. Meta-path-based neighbors can leverage structural differences in heterogeneous graphs to suppress edges based on meta-paths, disrupting short-range semantic connections between nodes. By multiplying a sequence of adjacency matrices, information from meta-path neighbors is acquired, compelling the model to explore alternative areas to predict masked relationships. Specifically, for a heterogeneous graph $G = \{ V,E\psi \}$, we generate a meta-path-based adjacency matrix ${A^{\psi} }$ for each meta-path $\psi  \in \Psi $ through meta-path sampling. To individually consider different meta-paths, we adopt the approach of masking and reconstructing each meta-path-based adjacency matrix separately. This is due to distinct meta-paths containing varied semantic information. For each ${A^{\psi} }$, we create a binary mask following a Bernoulli distribution $M_A^\psi  \sim pe$, with $pe < 1$ representing the masking rate of edges. Subsequently, we use $M_A^\psi $ to derive the adjacency matrix $\widetilde{{A^\psi }} = M_A^\psi  \cdot {A^\psi }$ based on the masked element path. Following this, we input the $\widetilde{{A^\psi }}$-associated node attributes $X$ to the encoder ${f_E}$ to obtain the latent node embedding $H_1^\psi $. The following steps are taken:

\begin{equation}
\label{eq_pe}
H_1^\psi  = {f_E}(\widetilde{{A^\psi }} ,X)
\end{equation}

We subsequently input $H_1^\psi $ nto the graph convolutional decoder ${f_D}$ to acquire the decoded node embeddings $H_2^\psi$. Subsequently,   is employed to reconstruct the adjacency matrix. The process is as follows:

\begin{equation}
	\label{eq_pe}
	H_2^\psi  = {f_D}(\widetilde {{A^{\psi}}},H_1^\psi ),{\kern 1pt} {\kern 1pt} {\kern 1pt} {\kern 1pt} {A^{{\psi ^{'}}}} = sigmoid({(H_2^\psi )^{T}} \cdot H_2^\psi ),
\end{equation}

Here, ${A^{{\psi ^{'}}}}$ represents the adjacency matrix reconstructed based on the meta-path after encoding and decoding. Subsequently, we compare the target ${A^\psi }$ and the feature edge reconstruction ${A^{{\psi ^{'}}}}$ for each node, utilizing the scaled cosine error as a metric.

\begin{equation}
	{{\rm I}^\psi } = \frac{1}{{\left| {{A^\psi }} \right|}}\sum\limits_{t \in {\rm T}} {(1 - \frac{{A_t^\psi  \cdot A_t^{{\psi ^{'}}}}}{{\left\| {A_t^\psi } \right\| \times \left\| {A_t^{{\psi ^{'}}}} \right\|}})} \tau 
\end{equation}

Where ${{\rm I}^\psi }$ represents the loss value of the meta-path $\psi $, and $\tau $ is the scaling scale ratio. We use the semantic-level attention vector $q$ to capture the crucial feature components ${{\rm P}^\psi }$ in each $\psi $. Next, the Softmax function is applied to normalize the cumulative score of all meta-paths, obtaining the weight information ${H^\psi }$ of $\psi $. The process is outlined below:

\begin{equation}
{{\rm P}^\psi } = {q^T} \cdot \tanh (W \cdot H_1^\psi  + b),{\kern 1pt} {\kern 1pt} {\kern 1pt} {H^\psi } = \frac{{\exp ({{\rm P}^\psi })}}{{\sum\nolimits_{\psi  \in \Psi } {\exp ({{\rm P}^\psi })} }}
\end{equation}

In this context, $W$ and $b$ represent the feature weight and feature bias, respectively. Ultimately, we compute the overall meta-path loss, deriving the comprehensive loss for meta-path-based edge reconstruction.

\begin{equation}
{{\rm I}_{MER}} = \sum\limits_{\psi  \in \Psi } {{{\rm Z}^\psi }}  \cdot {{\rm I}^\psi }
\end{equation}

\subsection{Convolutional masked autoencoder}
To mitigate the impact on prediction accuracy due to missing features in the modal state, we draw inspiration from the potent noise reduction capabilities of MAE. In lieu of the traditional MAE, we propose a CMAE to enhance the model's robustness, as illustrated in Figure.~\ref{fig:fram}($c$). CMAE comprises encoder and decoder components, both adopting ConvNeXt-based network structures. Through experimentation, we discovered that the MAE built on the transformer framework adeptly processes sequence information, demonstrating robust modeling capabilities. This approach allows a focus on the patch itself for acquiring rich semantic information and concurrently reducing pre-training costs. It's worth noting that the design of the transformer conflicts with ConvNets utilizing dense sliding windows. In general, ConvNets tend to copy-paste information from masked regions during feature extraction, whereas the transformer can use patches as the exclusive input to the encoder.

Motivated by sparse convolution, we seek to interpret the input within the masked area as a sparse data representation to address this issue \cite{choy20194d,xie2020pointcontrast}. The occluded graphical representation can be portrayed in experiments as a two-dimensional sparse array. To enhance the performance of masked autoencoder pre-training, we integrate sparse convolutions into CMAE. In the pre-training phase, standard convolutional layers in the encoder of ConvNets are transformed into sub-manifold sparse convolutions, enabling the model to operate solely on visible data points \cite{riasatian2021fine,choy20194d}. Consequently, CMAE leverages sparse convolution to implement ConvNeXt and seamlessly reverts to a standard convolutional layer during fine-tuning without any specialized processing of the sparse convolutional layer. Moreover, the same outcome can be achieved by employing binary mask operations before and after the convolution operation.

To enhance pre-training efficiency, we employ lightweight ConvNeXt blocks as the decoder of CMAE, rendering the entire design fully convolutional. Drawing inspiration from the optimization strategy of Swim-Transformer, ConvNeXt switches the optimizer from SGD to AdamW and incorporates regularization techniques like random depth, label smoothing, EMA, etc. Comprising multiple stages, each with several blocks, ConvNeXt adjusts the ratio to 1:1:3:1. Experimental findings indicate that layer normal (LN) outperforms batch normalization (BN) in ConvNeXt, prompting the use of LN in each block for model optimization. Importantly, we introduce global response normalization (GRN) technology to ConvNeXt, enhancing channel contrast and selectivity \cite{woo2023convnext}. In experiments, GRN accomplishes global feature aggregation, normalization, and calibration. We utilize the global function ${\rm O}( \cdot )$ to process the heterogeneous graph $H \in {R^{h \times w \times c}}$ transformed by feature edge reconstruction into. The process is outlined as follows:
\begin{equation}
{\rm O}(H): = H \in {R^{h \times w \times c}} \to og \in {R^c}
\end{equation}

The operation described above can be considered a pooling process. However, excessive use of pooling operations may result in a loss of details and information. Therefore, our utilization of norm-based feature aggregation, specifically employing ${L_2}$-norm, demonstrates improved performance. Assuming we have a set of aggregated values ${\rm O}(H) = og = \{ \left\| {{H_1}} \right\|,\left\| {{H_2}} \right\|,...,\left\| {{H_c}} \right\|\}  \in {R^c}$, we apply the response normalization function ${\rm N}( \cdot )$ to it. The process is:
\begin{equation}
{\rm N}(\left\| {{H_i}} \right\|): = \left\| {{H_i}} \right\| \in R \to \frac{{\left\| {{H_i}} \right\|}}{{\sum\nolimits_{j = 1,...,c} {\left\| {{H_j}} \right\|} }} \in R,
\end{equation}

Where ${G_i}$ is the $i$-th channel of ${L_2}$-norm, and formula (7) is equally crucial for calculating features in other channels. This operation induces feature competition within the channel through mutual inhibition. Finally, we utilize feature correction parameters to adjust the original input features. The process is::
\begin{equation}
H_i^{'} = {H_i} * {\rm N}({\rm O}{(H)_i}) \in {R^{h \times w}}
\end{equation}

\subsection{Feature cross fusion}
Following CMAE processing, our objective is to enhance the interaction among feature information within multimodal modes. This aims to enrich information expression, learn complementary features, and mitigate noise and deletions. To facilitate the interaction of multimodal features, we introduce a feature cross-fusion (FEC) module to establish communication between different modalities. This module primarily comprises two MLP layers to integrate features from CMAE. Assuming the input feature is $H \in {R^{B \times M \times C}}$, the process is:
\begin{equation}
\begin{array}{l}
	{H^{{\mathop{\rm int}} er}}: = {H^{'}}^T + ML{P_2}({(LN({H^{'}}))^T})\\
	{H^{{\mathop{\rm int}} er}}: = {({H^{{\mathop{\rm int}} er}})^{T}} + ML{P_3}({(LN({H^{{\mathop{\rm int}} er}}))^{T}}),
\end{array}
\end{equation}

Among them, the input is normalized using and contains two fully connected layers for feature interaction. After processing by the FEC module, information from different modalities is fully communicated to ensure that multimodal interaction contains information from multiple modalities.

Secondly, we jump-connect the feature map $H \in {R^{B \times 1 \times C}}$ reconstructed by the feature edge to the corresponding output feature ${H^{{\mathop{\rm int}} er}} \in {R^{B \times N \times C}}$ to obtain the multimodal mixed output $H^{"}$, which can be expressed as:
\begin{equation}
{H^{"}} = H \oplus {H^{{\mathop{\rm int}} er}}
\end{equation}

Which $ \oplus $ represents the addition operation of features, and $H^{"}$ contains all the information of this modality and the feature interaction information of other modalities.

\subsection{Strategies for model prediction of patient survival}
Considering the association of multimodal data with survival information, we investigate survival prediction in single-modal, two-modal, and multimodal scenarios. Thus, we aim to predict survival in seven different ways, as illustrated in Figure.~\ref{fig:fram}($d$). All multimodal data related to survival prediction are ultimately processed through an MLP, and the procedure is as follows:
\begin{equation}
W = MLP({H^{"}})
\end{equation}

Therefore, the calculation process of Cox loss for each mode   is as follows:

\begin{equation}
{\rm I}_{Cox}^m = \sum\limits_{i = 1} {{\delta _i}} ( - H_m^(i) + \log \sum\limits_{j,{t_j} \ge {t_i}} {\exp ( - H_m^(j))} )
\end{equation}

Among them, $H_m^{"}(i)$ and $H_m^{"}(j)$ represent the survival output of the  $i$-th and  $j$-th patients respectively. Therefore, the total loss of this framework is:
\begin{equation}
{{\rm I}_{total}} = {{\rm I}_{MER}} + \alpha {{\rm I}_{CMAE}} + \sum\limits_{m \in M} {\beta {\rm I}_{Cox}^m} 
\end{equation}

where $\alpha $ and $\beta$ represent the respective adjusted parameters. Considering the varied complexities of different modalities, we determine $\alpha  = 5$ and $\beta  = 1$ through extensive experimentation. In the framework's reasoning process, we initially conduct feature edge reconstruction on heterogeneous graphs to acquire semantic information enriched with edges. Subsequently, CMAE is employed to deduce features when partial modal data is available. Finally, the feature interaction module facilitates interaction among features within and across modalities. Our model demonstrates the capability to predict the survival of cancer patients in scenarios involving missing modal data and incomplete modalities.

Throughout the reasoning process, collaborative decision-making occurs among all functional modules, culminating in the survival prediction for cancer patients. To elucidate the data processing mechanism of SELECTOR, we employ pseudocode to intricately describe the model's inference process, as outlined in Algorithm 1.

\begin{algorithm}[!t]
	\small
	\caption{Model Inference.}\label{alg:alg1}
	\begin{algorithmic}
		\STATE
		\STATE \textbf{Input:} Multimodal data of ${M_p},{M_g},{M_c}$, Multimodal diagram $\{ {G_1},...,{G_m}\} ,m \in M$.		
		\STATE \textbf{Output:} Final survival prediction ${S_{final}}$.
		\STATE \textbf{Initialization:} Randomly initialize SELECTOR.
		\STATE \textcolor{blue}{\# Heterogeneous graph after feature edge reconstruction.}
		\STATE  \textbf{for} $m$ in ${M^{'}}$ \textbf{do}
		\STATE \hspace{0.5cm} ${G_m} = \sum\nolimits_{i = 1}^N {sigmoid({{({f_D}(\widetilde {{A^\psi }} ,H_i^\psi ))}^T} \cdot H_{i + 1}^\psi )} .$
		\STATE  \textbf{end for}
		\STATE \textcolor{blue}{\# Feature extraction of CMAE.}
		\STATE  \textbf{for} $m$ in ${M^{'}}$ \textbf{do}
		\STATE \hspace{0.5cm} ${H_m} = f_D^{CMAE}(f_E^{CMAE}({G_m}),{\rm N}(\left\| {{G_m}} \right\|)).$
		\STATE  \textbf{end for}

		\STATE \textcolor{blue}{\# Compute multimodal feature cross fusion.}
		\STATE $\begin{array}{l}
				{H^{{\mathop{\rm int}} er}}: = {H^T} + ML{P_2}({(LayerNorm(H))^T})\\
				{H^{{\mathop{\rm int}} er}}: = {({H^{{\mathop{\rm int}} er}})^T} + ML{P_3}({(LayerNorm({H^{{\mathop{\rm int}} er}}))^T}),
				\end{array}$ 
	
		\STATE \textcolor{blue}{\# Skip connection communication of intra-modal features.}
		\STATE  \textbf{for} $m$ in ${M^{'}}$ \textbf{do}
		\STATE \hspace{0.5cm} ${H^{"}} = {H^{'}} \oplus {H^{{\mathop{\rm int}} er}}.$
		\STATE  \textbf{end for}
		
		\STATE \textcolor{blue}{\# Compute predicted survival for available modalities.}
		\STATE  \textbf{for} $m$ in ${M^{'}}$ \textbf{do}
		\STATE \hspace{0.5cm} $W = MLP({H^{"}}).$
		\STATE  \textbf{end for}
		\STATE \textcolor{blue}{\# Multimodal fusion strategy}
		\STATE ${S_{final}} \leftarrow Mean({S_1},...,{S_2})$
		
	\end{algorithmic}
	\label{alg1}
\end{algorithm}

\section{Experiments}
\subsection{Datasets}
To assess the model's performance, experiments were conducted on datasets of kidney renal clear cell carcinoma (KIRC) (385 cases), liver hepatocellular carcinoma (LIHC) (287 cases), esophageal cancer (ESCA) (153 cases), lung squamous cell carcinoma (LUSC) (438 cases), lung adenocarcinoma (LUAD) (452 cases), and endometrial cancer (UCEC) (387 cases) \cite{tomczak2015review}. Our dataset comprises diagnostic pathology image data from TCGA, genomic data indicating censorship status, and clinical records. For pathology data processing, pre-trained models were employed to extract features from diagnosed WSIs. To ensure consistent resolution of local features, only WSIs with an average size of 20840 15606 were selected, and all slices were normalized to a $\times$10 magnification. Genomic data processing involved using the MSigDB gene set to filter genomic data through GSE \cite{reimand2019pathway}. The experiment ultimately selected five gene types (Tumor Suppression, Oncogenesis, Protein Kinases, Cellular Differentiation, Cytokines, and Growth), totaling 1,742 genes. For clinical record data processing, only patient age, gender, BMI, and other relevant information were selected.

\subsection{Experiments details}
The model is developed using Python 3.9 based on the PyTorch 1.13.1 platform. Both the model proposed in the paper and the compared model are trained using an NVIDIA RTX 4090 GPU. The hyperparameters of the model can directly affect its performance and the results of training. To determine the best hyperparameters, six primary parameters were mainly adjusted in the experiment: batch size, epoch, optimizer, learning rate, dropout, activation function, and weight decay, as shown in Table~\ref{tab:hyper}. 

\begin{table}[H]
	\caption{Selection range and optimal values of model hyperparameters} 
	\label{tab:hyper}
	\renewcommand{\arraystretch}{1}
	\centering
	\resizebox{1\linewidth}{!}{
	\begin{tabular}{c|c|c}
		\hline
		\textbf{Hyper-parameters} & \textbf{Select range} & \textbf{Optimal value}\\
		\hline
		batch size           & {64,128,256}        & 128\\
		epochs               & {100, 500, 1000}    &500\\
		optimizer            & {SGD, Adam, RMSProp}&Adam\\
		learning rate        & {3e-2, 3e-3, 3e-4}  & 3e-4\\
		Dropout              & (0.1,0.3,0.5)       & 0.3\\
		activation function  & (ReLu,Prelu)        & Prelu\\
		\hline
	\end{tabular}}
\end{table}

\subsection{Evaluation index}
This experiment employs the concordance index (C-index) and Kaplan-Meier (KM) analysis methods to assess the performance and prediction accuracy of the survival analysis model \cite{kattan2002postoperative}. The C-index ranges from 0 to 1, with a higher value indicating better prediction results. KM analysis, a commonly used survival analysis method, estimates the probability of observing an event (e.g., survival, recurrence, or death) at different time points and compares survival curves among different groups. For each dataset, it assesses patient stratification by dividing them into high-risk and low-risk groups based on the median score of the prediction model. Using the Log-Rank test, a lower $P$-value signifies better performance \cite{wang2019machine}. In the experiment, we employed the five-fold cross-validation method to train the model for predicting patient survival. Each TCGA dataset's samples were divided into five groups: four groups served as training sets, and one group as the test set. Additionally, during the training, 25\% of the samples from the training set were allocated to the validation set to fine-tune the model's weights. The model's five-fold cross-validation results were averaged to produce the final prediction.

\section{Results}
\subsection{Survival prediction of cancer patients}
We trained 13 models for comparisons in both unimodal and multimodal scenarios to evaluate their performance. Specifically, in the unimodal comparison, we included the Cox model based on clinical records, the Cox model based on genomic maps, the self-normalizing network (SNN), the improved XGBoost model (XGBLC), and the multi-instance learning (MIL) model based on WSI \cite{chen2022pan,matsuo2019survival,ma2022xgblc}. For multimodal performance comparison, we employed state-of-the-art methods, namely GSCNN \cite{mobadersany2018predicting}, MultiSurv \cite{vale2021long}, Pathomic \cite{chen2020pathomic}, Metric learning \cite{cheerla2019deep}, MCAT \cite{chen2021multimodal}, and HGCN. We conducted comprehensive ablation experiments on the proposed feature edge reconstruction and convolutional mask auto-encoder (CMAE) modules for both unimodal and multimodal scenarios.

As indicated in Table~\ref{tab:SOTA}, the MIL and Cox models in unimodal mode achieved performance mostly below 0.69 on WSI, genome mapping, and clinical information, respectively. When compared to the survival models' predictions in multimodality, we observed that the C-index values of GSCNN, MultiSurv, Pathomic, Metric learning, and MCAT were generally higher than the predicted values of the models under unimodality. This could be attributed to the fact that multimodal data enhance the diversity and richness of information, contributing to the survival prediction of cancer patients. Notably, our proposed SELECTOR model achieved C-index values of 0.768, 0.789, 0.699, 0.599, 0.665, and 0.78 on the six datasets of KIRC, LIHC, ESCA, LUSC, LUAD, and UCEC, respectively, surpassing the compared models significantly. This superiority is attributed to the inclusion of our proposed edge reconstruction module and convolutional mask autoencoder module in the SELECTOR model, primarily aimed at enhancing the prediction accuracy of the survival model. 

\begin{table*}
	\caption{Under unimodal and multimodal datasets, the 10 survival models obtain the most advanced C-index statistics, where bold indicates the best performance.} 
	\label{tab:SOTA}
	\renewcommand\arraystretch{1.5}
	\centering
	\begin{tabular}{cccccccc}
		\Xhline{1.5pt}
		Type & Methods & KIRC & LIHC &ESCA & LUSC&LUAD& UCEC\\
		\hline
		\multirow{4}*{Unimodal}& Cox($c$, 2019) \cite{matsuo2019survival} & 0.689 & 0.506 &0.591 &0.567 &0.567& 0.673\\
		& Cox($g$, 2019) \cite{matsuo2019survival}  & 0.671 & 0.657 &0.518 &0.524 &0.621 & 0.680\\
		& MIL($p$, 2022) \cite{chen2022pan} & 0.661 & 0.642 &0.618 &0.561 &0.567 &0.670 \\
		& SNN($c$, 2022) \cite{chen2022pan} & 0.633 & 0.594 & 0.626&0.522 &0.554 &0.580 \\
		& XGBLC ($c$, 2022) \cite{ma2022xgblc} & 0.716 & 0.703 & 0.642&0.599 &0.673 &0.712 \\
		\hline
		\multirow{8}*{Multimodal} & GSCNN(2018) \cite{mobadersany2018predicting} & 0.713 & 0.655& 0.550 & 0.546 & 0.617 & 0.726\\
		& Metric learning(2019) \cite{cheerla2019deep} & 0.729 & 0.668 &0.604 &0.582 &0.613 & 0.690\\
		& MultiSurv(2021) \cite{vale2021long}  & 0.676 & 0.623 &0.586 &0.564 &0.626 &0.690 \\
		
		& MCAT(2021) \cite{chen2021multimodal} & 0.672 & 0.685 & 0.576&0.564 &0.608 &0.683 \\
		& MMF(2022) \cite{chen2022pan} & 0.659 & 0.622 & 0.613 &0.538 &0.600 &0.634 \\
		& Pathomic(2022) \cite{chen2020pathomic} & 0.691 & 0.662 &0.602 &0.560 &0.602 & 0.676\\

		& HGCN(2023) \cite{hou2023hybrid} & 0.747 & 0.693 & 0.634 &0.598 &0.651 &0.747 \\
		& SELECTOR & 0.768 & 0.789 & 0.699&0.599 &0.665 &0.780 \\
		\Xhline{1.5pt}
	\end{tabular}
\end{table*}

\begin{figure*}[!t]
	\centerline{\includegraphics[width=1.5\columnwidth]{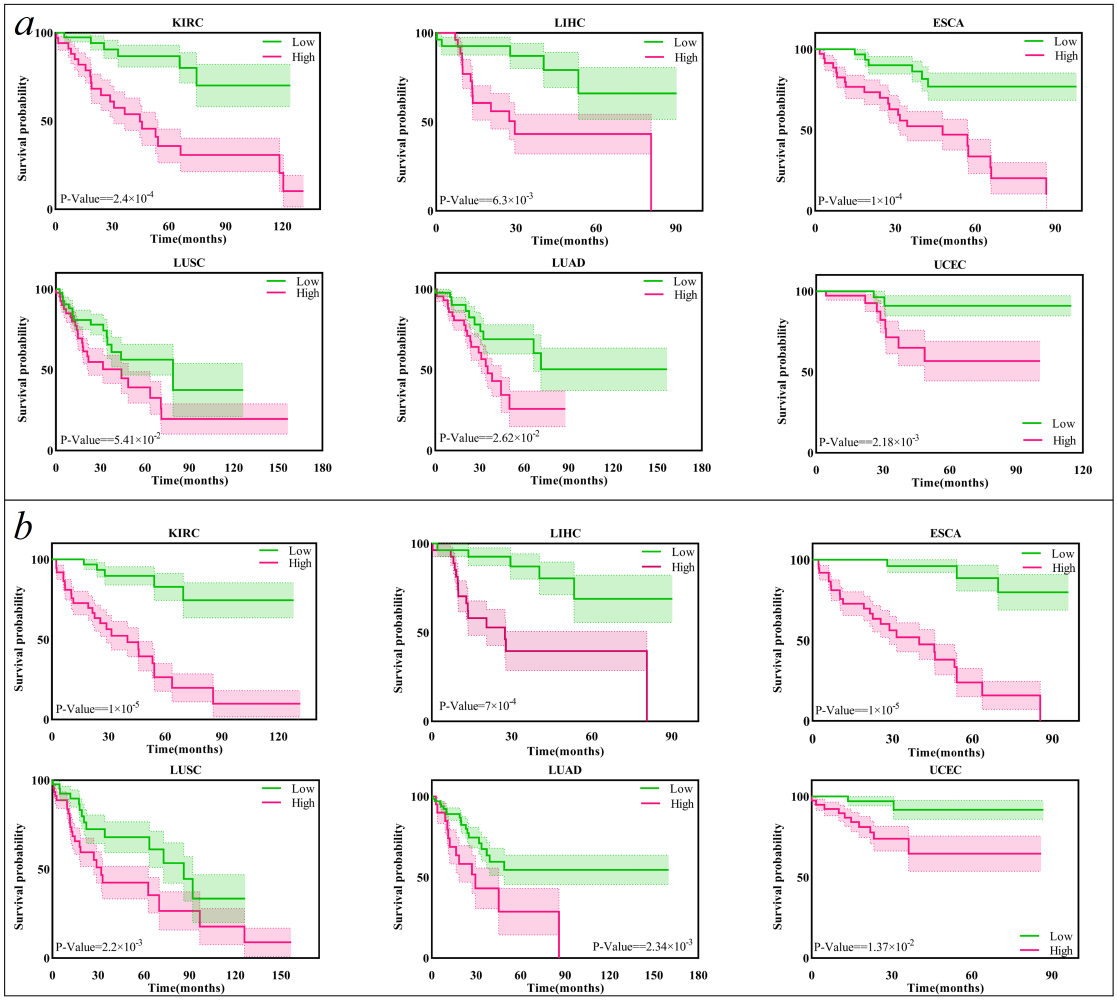}}
	\caption{Comparison of the optimal method of KM analysis and SELECTOR in six cancer datasets. For each dataset, patients were divided into high-risk groups (red zone) and low-risk groups (green zone) based on the median score output of the prediction model. The $P$ value for each Log-rank test is placed in the corner of each figure. $a$ shows the best model (HGCN) to obtain survival predictions. $b$ shows the survival prediction obtained by SELECTOR.}
	\label{fig:survival}
\end{figure*}

In the test set, we utilized the model's output to generate median survival risk values, enabling the classification of patients in each cancer dataset into high and low risk groups. For constructing the survival curve, the Log-Rank test was employed to discern significant differences between two or more survival curves. This statistical test calculates a statistic based on the observed number of event occurrences and the risk set at each time point, comparing it with the theoretical expected curve \cite{miyamoto2015sarcopenia}. A significant difference is considered when the $p$-value of the statistic is less than a preset significance level (usually 0.05) \cite{xie2005adjusted}. Figure.~\ref{fig:survival}($a$) illustrates the survival prediction map obtained by the best method, where HGCN serves as our baseline with the best prediction effect. From Figure.~\ref{fig:survival}($b$), it is evident that SELECTOR outperforms others in KIRC, LIHC, ESCA, LUSC, and LUAD. The P values obtained on the six cancer datasets of UCEC were 0.00024, 0.0007, 0.0002, 0.0022, 0.00234, and 0.00218, respectively. The P value is smaller than other methods, and Figure.~\ref{fig:survival}($b$) demonstrates that SELECTOR can significantly distinguish high- and low-risk patients in all cancer datasets. Therefore, SELECTOR stands out as the best method for predicting the survival of cancer patients.

\subsection{Interpretability of results}
\begin{figure*}[!t]
	\centerline{\includegraphics[width=2\columnwidth]{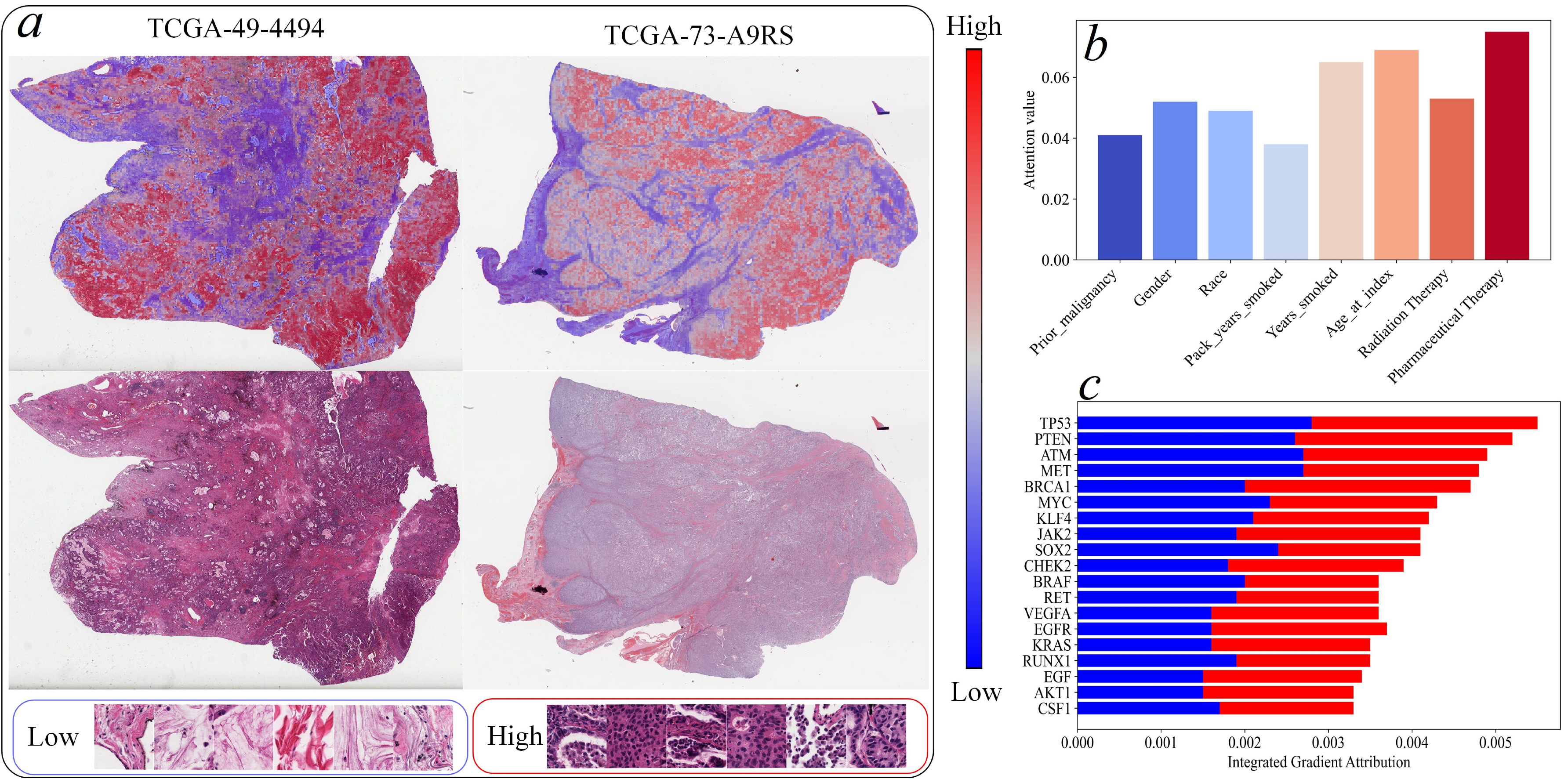}}
	\caption{Explanation of the SELECTOR framework on the LUAD dataset. $a$: Pathological section heat map. $b$: Concern value of clinical records. $c$: Comprehensive gradient analysis of genetic maps.}
	\label{fig:inter}
\end{figure*}

The survival model for cancer patients has traditionally been perceived as a black box. Nevertheless, in the medical field, the interpretability of survival models is crucial for aiding doctors and clinical decision-makers in comprehending patient survival probabilities and risks. It also contributes to personalized precision medicine. Leveraging the modeling of multimodal data, SELECTOR offers enhanced interpretability through the output of multimodal graphs and integrated gradient analysis. In Figure.~\ref{fig:inter} ($a$), the model's interpretability of cancer microenvironments in pathological images is visualized. The areas with high and low attention values in patches of high- and low-risk patients can be explored to discover new biomarkers. Figure.~\ref{fig:inter}($b$) illustrates the contribution values of sample clinical indicators. Notably, drug treatment and age are of significant interest in predicting patient survival with SELECTOR. Figure.~\ref{fig:inter}($c$) displays the gene map data, showcasing the distribution of gene feature values corresponding to patients in high- and low-risk groups, facilitating an understanding of the relative contribution of tumor molecules in the model.

\subsection{Ablation experiment}
\begin{figure*}[!t]
	\centerline{\includegraphics[width=2\columnwidth]{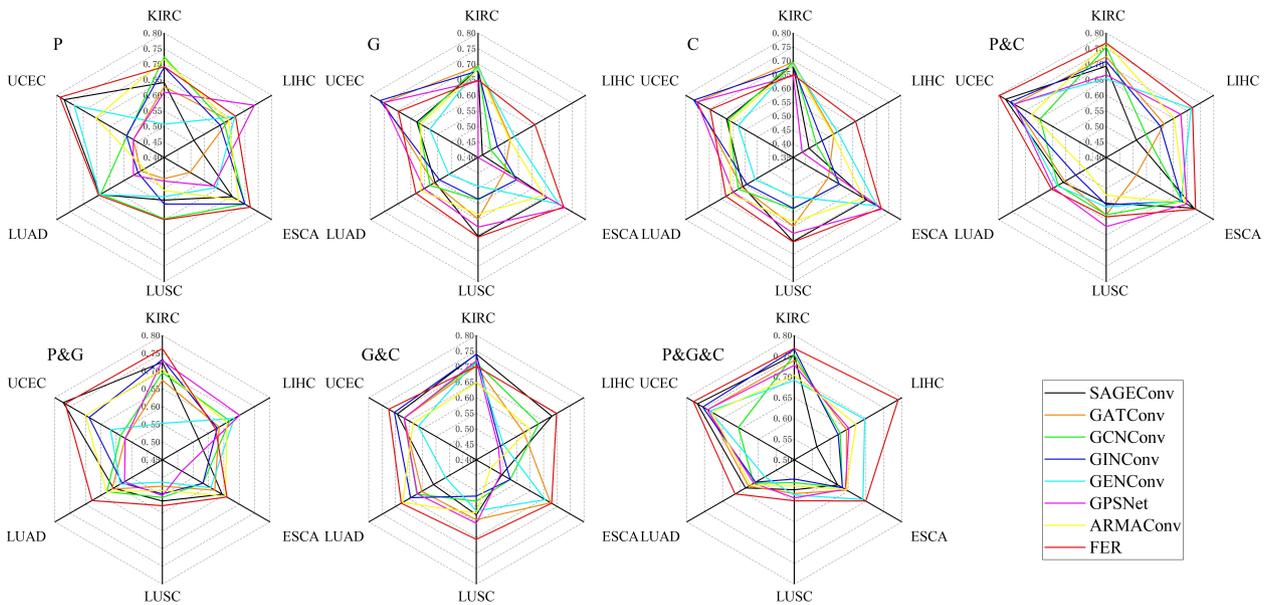}}
	\caption{Models containing different graph convolutions predict survival and obtain C-index radar plot of ablation experiment results. $P$ represents pathological modality, $G$ represents gene expression modality, $C$ represents clinical data, and \& represents a combination of modalities.}
	\label{fig:graph}
\end{figure*}

To assess the impact of the FER module on SELECTOR, we conducted tests with eight schemes in both unimodal and multimodal settings across six cancer datasets: KIRC, LIHC, ESCA, LUSC, LUAD, and UCEC, all of which involve graph convolutional paradigms. The paradigms include SAGEConv, GATConv, GCNConv, GINConv, GENConv, GPSNet, ARMAConv, and FER \cite{islam2020learning,li2020pooling,wang2022cell,benovoy2009ectopic,lin2020gps,wang2022spatiotemporal}. Figure.~\ref{fig:graph} presents the results of the FDR module ablation experiment. In $P$, $G$, and $C$ modes, SELECTOR with the FER module exhibited a favorable predictive effect, particularly on LIHC, ESCA, LUSC, and LUAD cancer datasets. Other module-containing survival prediction models achieved the best results for KIRC and UCEC. In dual modality, SELECTOR also demonstrated superior results in survival prediction across approximately five cancer datasets. In the multimoda setting, SELECTOR consistently outperformed other models across all cancer datasets. Additionally, from a modal perspective, the increased utilization of modal data features correlated with enhanced prediction performance of SELECTOR. This is attributed to multimodal providing more patient survival-related information. In summary, the C-index value of SELECTOR with the FER module is significantly higher than that of prediction models involving other graph networks. This is due to FER's ability to learn effective node embeddings in multimodal graphs and fully consider heterogeneous graphs. The complex structure contains semantic information, contributing to performance improvement.

\begin{figure*}[!t]
	\centerline{\includegraphics[width=2\columnwidth]{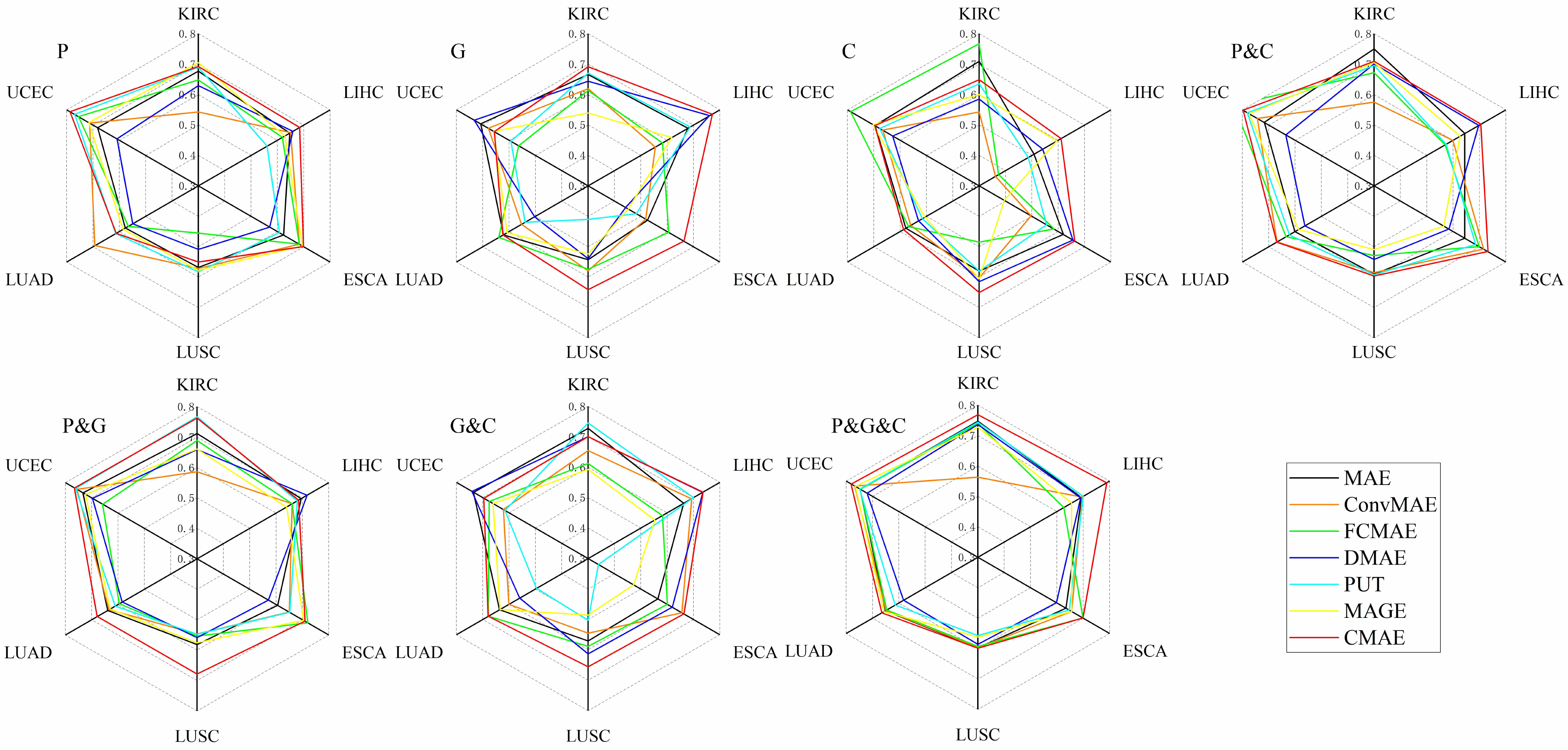}}
	\caption{Radar plot of ablation experiment results for C-index obtained by model prediction survival with different mask decoders. $P$ represents pathological modality, $G$ represents gene expression modality, $C$ represents clinical data, and \& represents a combination of modalities.}
	\label{fig:mae}
\end{figure*}

To assess the impact of the CMAE module on SELECTOR, we tested seven schemes in both unimodal and multimodal settings across six cancer datasets: KIRC, LIHC, ESCA, LUSC, LUAD, and UCEC, which also involve MAE paradigms. These include MAE, ConvMAE, FCMAE, DMAE, PUT, MAGE, and CMAE \cite{he2022masked,gao2022convmae,woo2023convnext,wu2022denoising,liu2022reduce,li2023mage}. Figure.~\ref{fig:mae} displays the results of the ablation experiment for the CMAE module. In $P$, $G$, and $C$ modes, SELECTOR with CMAE showed good predictive performance for KIRC, LIHC, and ESCA cancer datasets. The survival prediction model with MAE achieved the best results for LUSC, LUAD, and UCEC. In dual modality, SELECTOR also achieved the best results in survival prediction for approximately four cancer datasets. In the multimodal setting, SELECTOR consistently outperformed other models across all cancer datasets. We observed that the prediction performance of SELECTOR in the multimodal case is significantly superior to that in the unimodal setting, as unimodal lacks the rich semantic information of multimodal. Additionally, SELECTOR yielded poor results for some cancer datasets in both unimodality and bimodality, possibly due to insufficient data volume leading to model overfitting or improper parameter tuning. In summary, the C-index value obtained by SELECTOR with the CMAE module for predicting survival is significantly higher than that of survival models with other MAEs. This is because CMAE can robustly learn all the information in multimodal data without being affected by data missing and interference from data errors.

\section{Discussion}
\subsection{Analysis}
With medical imaging, electronic health records, and omics data accumulating at an unprecedented rate, artificial intelligence algorithms in cancer prognosis represent a research area of great potential. Multimodal medical data contains more comprehensive and richer semantic patient information, enhancing the analysis and decision-making capabilities of survival models. However, multimodal data is prone to modal data loss and recording errors, posing serious problems and challenges to automated analysis methods. This paper proposes constructing a heterogeneous graph for multimodal data to represent various types of nodes and edges, facilitating the flexible capture of complex relationships in different modal data. Additionally, the multimodal heterogeneous graph we constructed allows each node and edge to carry different types of information, improving the understanding of relationships between nodes.

Leveraging the construction of heterogeneous graphs, we propose SELECTOR, a heterogeneous graph-aware network utilizing convolutional mask encoders to extract features from multimodal data. Employing the FER module, we optimize the heterogeneous graph to address missing edge information and mitigate the impact of noise. The reconstruction of edges introduces additional relevant information, enhancing the richness of the heterogeneous graph data and improving data visualization. Through an analysis of the ablation experimental results of the FER module, it is evident that the reconstruction of edge information plays a significant role in enhancing the performance of SELECTOR for survival prediction.

Additionally, for enhanced model robustness, we introduce a convolutional mask encoder to extract reconstructed feature maps. The sparse convolutional layer selectively captures discernible features in CMAE, thereby reducing computation and parameters, leading to improved computational efficiency and generalization. The convolution operation facilitates acquiring the local receptive field of heterogeneous graph features, aiding in capturing multi-scale information from the data. In comparison to transformer-based MAE, CMAE achieves reduced computation and robust learning of heterogeneous graph features. Analysis of the ablation experimental results of the CMAE module indicates that the masked convolutional encoder significantly enhances the survival prediction performance of SELECTOR. However, this enhancement comes at the cost of increased model parameters and equipment overhead.

To enhance the interaction with feature information across different modalities, enrich information expression, learn complementary features, and address noise and missing data issues, we introduce a FEC module. This module is designed to establish communication between various modalities. In the model's inference phase, we utilize six patterns to predict cancer patient survival. SELECTOR can provide survival predictions for cancer patients even when certain modalities or data within a modality are missing. Extensive experiments were conducted on six cancer datasets from TCGA, evaluating the performance and prediction effectiveness of SELECTOR based on C-index and survival prediction curves. Additionally, we interpret SELECTOR through model input and output. Furthermore, to assess the impact of the FEC and CMAE modules, comprehensive ablation experiments were performed on six cancer datasets from TCGA. SELECTOR demonstrated superior performance in both multimodal and unimodal scenarios.

We performed comprehensive experiments on six cancer datasets from TCGA, assessing performance and prediction effectiveness using C-index and survival prediction curves to verify the capability of SELECTOR. Additionally, we conducted interpretable analysis of SELECTOR through examination of model input and output. Subsequently, to validate the impact of the FEC and CMAE modules, we conducted extensive ablation experiments on six cancer datasets from TCGA. SELECTOR demonstrated superior performance in both multimodal and unimodal conditions.

\subsection{Limitation}
Some limitations of this study need to be addressed. The selection of an appropriate feature fusion method is crucial for survival prediction modeling. Spliced feature vectors enable interaction between features of different modalities, capturing more complex correlations. However, the feature fusion method employed is simple splicing, potentially treating all features equally and overlooking the importance of features from different modalities, resulting in potential underestimation or overestimation of features from specific modalities. Second, the issue of labeling imbalance is more pronounced due to the limited dataset of cancer in TCGA and some data loss during precision, leading to an imbalance of survival labels in the model. Lastly, the challenge of model interpretability for multimodal data must be acknowledged. The substantial number of parameters and complex structure in SELECTOR complicates the internal arithmetic process, reducing interpretability. Model interpretability is further influenced by biases and tendencies inherent in the data. In this paper, we solely employ the five-fold cross-validation method to assess the model's performance.

\subsection{Prospect}
Predicting cancer survival time is a crucial research area with extensive applications in medicine. By thoroughly analyzing patients' gene expression, imaging, and clinical data, we can achieve more accurate predictions of patients' survival times. This provides a critical foundation for doctors to devise treatment plans and make decisions regarding survival extension strategies. Additionally, doctors can leverage the predicted results to identify patients in need of more aggressive treatment and those who may benefit from specific drugs or therapies, facilitating personalized and precise treatment strategies. Cancer survival prediction contributes to early screening and intervention for high- and low-risk groups, ultimately enhancing the survival rates and treatment outcomes for cancer patients.

\section{Conclusion}
In this study, we introduce SELECTOR, a heterogeneous graph-aware network utilizing a convolutional mask encoder for robust multimodal prediction of cancer patient survival. Initially, we construct a multimodal heterogeneous graph and employ the meta-path method for feature edge reconstruction. This process ensures a comprehensive consideration of feature information from graph edges and the effective embedding of nodes. To address the impact of missing features within the modality on prediction accuracy, we introduce CMAE, which processes the heterogeneous graph after feature reconstruction. CMAE primarily utilizes sparse convolution to extract only unmasked features during this process. Additionally, the FEC module is implemented to establish communication between modalities, ensuring that the output features encompass all modality-specific features along with relevant information from other modalities. Our extensive experiments on six cancer datasets confirm the validity and reliability of SELECTOR, especially in scenarios involving missing modalities and intra-modal information. In future research, we plan to focus on integrating valid information across multiple modalities and exploring new and efficient multimodal prediction methods.

\section*{CRediT authorship contribution statement}
Liangrui Pan, Yijun Peng, Yan Li: Conceptualization, Methodology, Writing – original draft, Writing – review \& editing. Xiang Wang, Wenjuan Liu, Liwen Xu: Conceptualization, Validation, Visualization. Qingchun Liang and Shaoliang Peng:Supervision.

\section*{Declaration of competing interest}
There are no funds and conflict of interest available for this manuscript.

\section*{Acknowledgment}
\begin{sloppypar}
This work was supported by NSFC-FDCT Grants 62361166662; National Key R\&D Program of China 2023YFC3503400, 2022YFC3400400; Key R\&D Program of Hunan Province 2023GK2004, 2023SK2059, 2023SK2060; Top 10 Technical Key Project in Hunan Province 2023GK1010; Key Technologies R\&D Program of Guangdong Province (2023B1111030004 to FFH). The Funds of State Key Laboratory of Chemo/Biosensing and Chemometrics, the National Supercomputing Center in Changsha (\url{http://nscc.hnu.edu.cn/}), and Peng Cheng Lab.
\end{sloppypar}


\bibliographystyle{unsrt}

\bibliography{REFERENCES}

\end{document}